\documentclass[runningheads]{llncs}
\usepackage{graphicx}
\usepackage{comment}
\usepackage{amsmath,amssymb} 
\usepackage{color}
\usepackage{mathrsfs}
\usepackage{commath}
\usepackage{comment}
\usepackage{soul,color}
\usepackage{subcaption}
\usepackage{multirow}

\usepackage[width=122mm,left=12mm,paperwidth=146mm,height=193mm,top=12mm,paperheight=217mm]{geometry}

\begin{document}
\pagestyle{headings}
\mainmatter
\def\ECCVSubNumber{3631}  

\title{Potential Field: Interpretable and Unified Representation for Trajectory Prediction} 

\titlerunning{Potential Field: Representation for Trajectory Prediction}
\author{Shan Su\inst{1,2} \and
Cheng Peng\inst{3} \and
Jianbo Shi\inst{1} \and
Chiho Choi\inst{2}}
\authorrunning{Shan Su et al.}
%
\institute{University of Pennsylvania \and
Honda Research Institute USA \and
University of Minnesota }
\maketitle

\begin{abstract}
Predicting an agent's future trajectory is a challenging task given the complicated stimuli (environmental/inertial/social) of motion. Prior works learn individual stimulus independently and fuse the representations in an end-to-end manner, which makes it hard to understand what are actually captured and how they are fused. In this work, we borrow the notion of potential field as an interpretable and unified representation to model all stimuli. This allows us to not only supervise the intermediate learning process, but also have a coherent method to fuse the information of different sources. From the generated potential fields, we further predict the direction and speed of future motion, which are modeled as Gaussian distributions to account for the multi-modal nature of the problem. The final prediction results are generated by recurrently moving past location based on the predicted motion. We show state-of-the-art results on the ETH, UCY, and Stanford Drone datasets. 
\end{abstract}

\section{Introduction}

Trajectory prediction is essential for the safe operation of vehicles and robots designed to cooperate with humans. Although intensive research has been conducted, accurate prediction of road agents’ future motions is still a challenging problem given the high complexity
of stimuli~\cite{1905.06113}. To properly model the behavior of humans, three types of stimuli should be considered: (i) Environmental (external) stimulus: humans obey the physical constraints of the environment as they move on the walkable area and avoid collision with stationary obstacles; (ii) Inertial (internal) stimulus: humans’ future motions are driven by their own intention inferred from the past motion\footnote{We aim at the prediction task where agents have intentions, other than crowd simulation setup where agents move in completely open area without specific intentions. }; and (iii) Social stimulus: humans interactively negotiate for possession of the shared physical environment.
Meanwhile, prediction of human behavior is inherently multi-modal in nature. Given the past motion, there exist multiple plausible future trajectories due to the unpredictability of the future.

\begin{figure}[t!]
\centering
    \includegraphics[width=\textwidth]{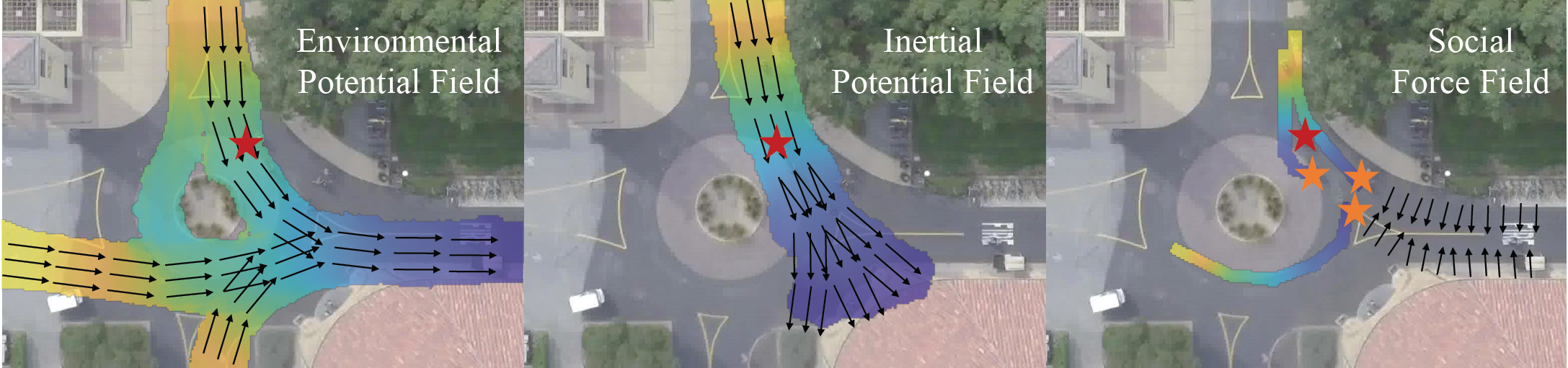}
    \caption{We address trajectory prediction problem from bird-eye view by introducing potential field. 
    Yellow and blue represent high and low potential values respectively and arrows indicate the motion/force direction. The target agent is marked in red while the neighbor agents are marked in orange. }
    \vspace*{-4mm}
    \label{fig:teaser}
\end{figure}

There have been research efforts~\cite{chandra2019traphic,lee2017desire,Sadeghian_2019_CVPR,Zhao_2019_CVPR} to model environmental, inertial, and social stimuli for trajectory prediction. They extract features of individual stimulus independently and fuse them in a feature space. Although such methods could be convenient to train the network in an end-to-end manner, 
the current models can not ensure whether the desired stimuli are actually captured (i.e., lack of interpretability) or whether the captured features are appropriate to fuse (i.e., lack of unified representation). Inspired by the Vygotsky’s zone of proximal development theory~\cite{chaiklin2003zone}, which claims the necessity of incremental supervision for learning tasks, we propose to supervise and evaluate the intermediate learning progress using \textbf{interpretable} representations that can be easily \textbf{unified} over different stimuli. 

\begin{figure*}[t!]
\centering
    \includegraphics[width=\textwidth]{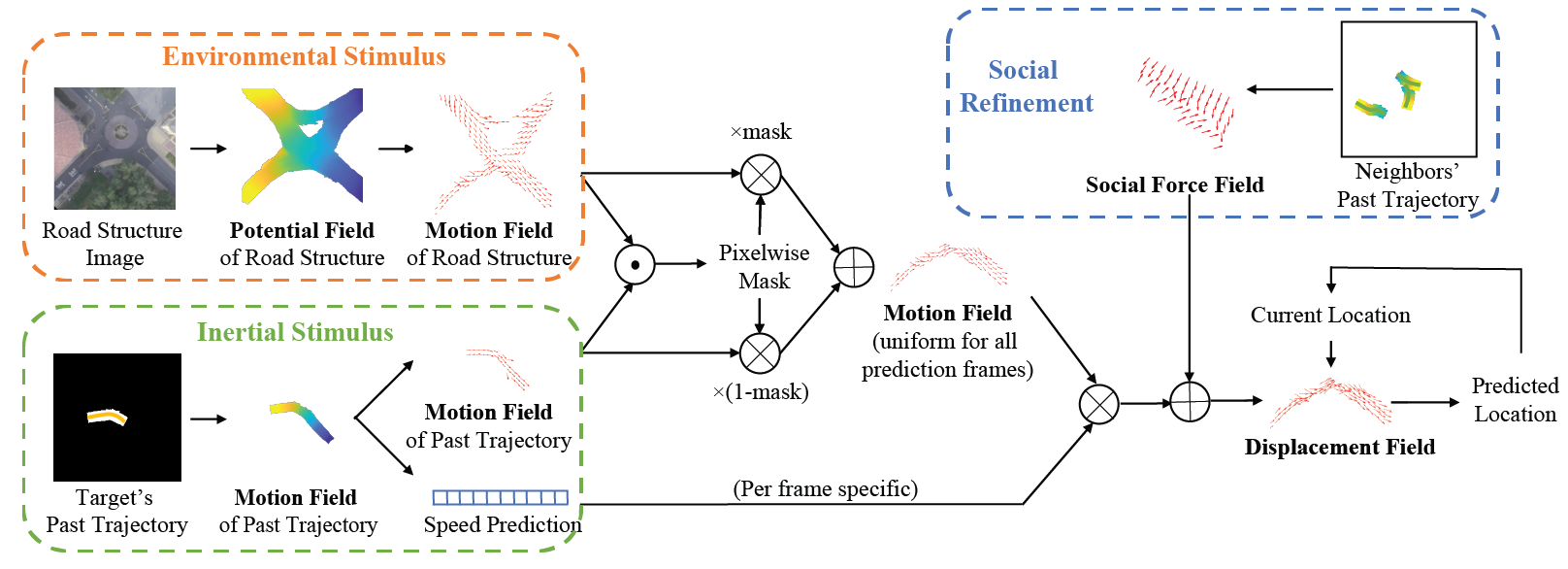}
    \caption{Overview of the proposed pipeline. 
    $\odot$ denotes concatenation, $\oplus$ denotes pixel-wise addition, and $\otimes$ donates multiplication.}
    \vspace*{-4mm}
    \label{fig:pipeline}
\end{figure*}

In this work, we present a novel way to model environmental, inertial, and social stimuli using \textit{invisible forces}, given the fact that force is the governing and unified factor of all interactions and motions. To account for the unpredictability of the problem (detailed in Section~\ref{sec:labeling}, Section~\ref{sec:prediction}), we borrow the notion of potential field as our primary representation. More specifically, we model the stimuli by environmental, inertial potential fields, and social force field as shown in Figure~\ref{fig:teaser}. Using potential field as an interpretable and unified representation, our method is able to supervise and seamlessly fuse the effect from three types of stimuli. Our conjecture is that such framework helps the network to comprehensively develop the intellectual capabilities~\cite{chaiklin2003zone}, and to model the human-level understanding by introducing domain knowledge support~\cite{lake2017building}.
The main contributions of this work are as follows:
(i) To the best of our knowledge, our method is the first to present potential field as a representation to model multiple stimuli for trajectory prediction. (ii) We develop a novel method to inversely formulate potential field from the observed trajectory and provide its mathematical derivation. 
(iii) We generalize our potential field knowledge to unknown scenes using a neural network. 
(iv) We develop a fully interpretable pipeline to do trajectory prediction from potential field with neural networks shown in Figure~\ref{fig:pipeline}.
(v) We achieve state-of-the-art performances on three widely used public datasets.  

Our proposed representation differs from traditional potential field concepts in the following ways. 
First, the field is generated from surrounding context automatically in a data driven way. 
Secondly, the generated field is time and pedestrian specific. 
Last but not the least, we introduce three fields corresponding to environmental, initial and social effect respectively. The three fields collectively determine the final prediction results even when one or more fields do not have enough constraints on human motions.

We organize the paper in the following way. In section~\ref{sec:labeling}, we present the method to inversely estimate potential values from observed trajectories. It is then served as the ground truth for training in the following sections. In section~\ref{stimuli_pf}, we propose environmental and inertial potential fields and train networks to generalize to unseen scenarios. In section~\ref{sec:prediction}, we propose to use the estimated potential fields, together with social force field, to predict pedestrian trajectories.

\section{Related Work} 

\noindent\textbf{Classic Models}
Classic models for future trajectory can be traced back to Newtonian mechanics, where future trajectory is calculated by force~\cite{griffiths2005introduction}. Such physics based models are further developed by introducing statistical models, such as Kalman filters~\cite{kalman1960new} and Gaussian process~\cite{williams1998prediction,wang2007gaussian}. Some approaches such as \cite{helbing1995social,luber2010people,zanlungo2011social} adopt a conceptual meaning of force to model attractive and repulsive behavior of humans. Classic models are revisited recently by~\cite{scholler2020constant}, where constant velocity model is used and achieves state-of-the-art performance. Such method has high interpretability by hand-crafting the motion feature and dynamics. However, it is not guaranteed that the hand-crafted dynamics rule actually models human motion in real life. Also, it is hard to incorporate contextual information, and thus limits the ability of extending the methods to complicated environments. In contrast, we target the problems which have complex context (\textit{e.g.}, road structure and interaction with other agents) and develop our method by combining the classic concepts with a data-driven approach. 

\noindent\textbf{Deep Learning Methods} 
Focusing on social interactions between humans, deep learning-based approaches for trajectory prediction have been studied to train the network to generate interaction-encoded features. A social pooling module is proposed in \cite{alahi2016social,gupta2018social,su2017predicting} where the embeddings of the agents' motion are aggregated in a feature space. They motivated the following works to capture internal stimuli using the additional input modality such as head pose~\cite{hasan2018mx} or body pose~\cite{yagi2018future} of humans. These approaches concatenate the feature representations of a pose proxy with motion embeddings, seeking to find a correlation between gaze direction and the person's destination. Besides these efforts, the structural layout of the road is commonly used to model external stimuli in~\cite{lee2017desire,xue2018ss,Zhao_2019_CVPR,Choi_2019_ICCV}. They first extract image features from RGB input and simply fuse with different types of feature embeddings. Although such models are convenient to use auxiliary information, they may not be optimal to capture desired stimuli and to fuse different feature embeddings~\cite{scholler2020constant}. Therefore, we introduce interpretable representations for intermediate supervision and their unified format to share domain knowledge.

\noindent\textbf{Potential Field}
Potential field methods are previously used for path planning in Robotics with an emphasis on obstacle avoidance in both static and dynamic environments~\cite{barraquand1992numerical,reif1999social,ge2002dynamic}. They use the potential field as a global representation of the space~\cite{hwang1992potential}. Such classic potential field method suffers from the methodology that values are heuristically assigned for both robots and obstacles. Choosing the hyper-parameters are tedious, and more importantly, the resulting trajectory may be sub-optimal~\cite{ge2002dynamic}. On the contrary, our method inversely estimate the field from surrounding context automatically in a data driven way. 

\section{Potential Value Labeling}\label{sec:labeling}
Force represents the overall effects on the target agent. However, force itself cannot account for the unpredictability of the future due to its deterministic direction and magnitude. We thus borrow the notion of potential field as our primary representation. 
In physics, a positive electric charge moves from high potential region to low potential region. In analogy to this, 
we define a potential field in traffic scenes, on which the agents can travel to anywhere with lower potential values. 

We introduce a potential field by assigning each location a scalar value that represents the potential energy. A pedestrian's motion is thus generated by moving towards locations with lower potential energy. 
Due to the fact that human motions do not have large acceleration or deceleration in everyday activities\footnote{It is also observed from the public datasets such as~\cite{pellegrini2009you,lerner2007crowds}.}, we assume that our invisible forces are proportional to \textit{velocities}, not the accelerations, of humans. 

In contrast to the heuristic method used in robotics, we aim to generate potential field in a data driven manner, so that hand-picking the appropriate hyper-parameters is not needed. This also guarantees that the generated potential values are compatible with actual human motions. However, there exists no ground-truth potential values for human motion, and it is impossible for humans to annotate such continuous and yet abstract labels by hand. In this section, we propose a way to inversely estimate potential values from the observed trajectories. 
In another word, our method can automatic label the given trajectory with potential values that is compatible with the observed motions.


\subsection{Derivation}\label{3_1}

We define a sequence of pedestrian's trajectory as a set of distinct points $X= \{x_1,...,x_T\}$, from time $1$ to $T$ with a constant sampling time interval of $\delta t$\footnote{In this paper, we set $\delta t = 0.4$ seconds.}. We use the notation $p(x)$ to denote the potential value of a point $x$ on the trajectory, and the notation $\mathcal{P}(u,v)$ to denote the potential value at a pixel $(u,v)$ in image coordinate $(\mathcal{U}, \mathcal{V})$. 

With an analogy to a positive electric charge's motion in electric field, the trajectory is modeled as movements of an agent towards lower potential value locations. It means that the potential values along a single trajectory should monotonically decrease. However, such decreasing property is not sufficient to generate a unique and stable field\footnote{The detailed comparison is shown in the supplementary material (A1).}. Therefore, 
we explicitly compute the potential values $p(x)$ for each point on the trajectory and infer a dense field in image coordinate from those sparse values. 
Our key observation is that the potential difference is linearly proportional to agents' velocity, which can be extracted from distance among points on the trajectory. It allows us to draw a direct relationship between \textit{distance} and \textit{potential values}. 

Given three adjacent points $x_i$, $x_{i+1}$, and $x_{i+2}$ on a trajectory $X$, their corresponding potential values are denoted as $p(x_i)$, $p(x_{i+1})$ and $p(x_{i+2})$. We assume that the velocity within a pair of adjacent sampled positions is constant. Therefore, the velocity within two points $(x_i, x_{i+1})$ is given as follows:
\begin{equation} \label{eq1}
    v_{i}=\frac{||x_{i+1}-x_i||_2}{\delta t}=\frac{d_{i}}{\delta t},
\end{equation}
where $\delta t$ is the sampling interval and $d_i$ is the distance between $x_i$ and $x_{i+1}$. Note that the velocity can be different for other segment regions in the same trajectory.

We denote the potential difference between two points $(x_i, x_{i+1})$ as $\delta p_{i} = p(x_i) - p(x_{i+1})$. Similar to the constant velocity assumption, we assume the derivative of the potential energy is constant from $x_i$ to $x_{i+1}$. The field strength is then denoted as $\mathcal{E}_{i} = \delta p_{i} / d_{i}$. 

In order to derive the relationship between the velocity $v$ and the potential difference $\delta p$, we borrow the potential field analogy from physics~\cite{griffiths2005introduction}. In theory of electricity, $\delta p$ is usually referred to as \textit{voltage} and $\mathcal{E}$ is referred to as \textit{electric field strength}. The corresponding electric force is then proportional to the electric field strength following $F=\mathcal{E}q$, where $q$ is the electric charge of an object. Similarly, we define our potential energy difference to be directly proportional to velocity $v$. Then, the velocity can be formulated as follows:
\begin{equation} \label{eq2}
    v_{i} = \alpha\mathcal{E}_{i} = \alpha\frac{p(x_{i})-p(x_{i+1})}{d_{i}},
\end{equation}
where $\alpha$ is a constant scaling factor that depends on the types and intrinsic properties of agents, which is similar to mass or electric charge of the objects in the theory of electricity. Note that the value $\alpha$ does not change throughout the same trajectory. By combining Eq.~\ref{eq1} and Eq.~\ref{eq2}, the relationship among potential values $p(x_i)$, $p(x_{i+1})$ and $p(x_{i+2})$ is derived as follows:
\begin{equation} \label{eq3}
    \frac{p(x_i)-p(x_{i+1})}{p(x_{i+1})-p(x_{i+2})} = \frac{d_{i}^2}{d_{i+1}^2}.
\end{equation}

The constant velocity and uniform field strength assumptions require three points to be adjacently sampled. 
We further generalize\footnote{The detailed derivations are provided in the supplementary material (A2).} Eq.~\ref{eq3} to the potential values among any triplets $(x_i, x_j, x_k)$ on the same trajectory as follows:
\begin{equation} \label{eq6}
    \frac{p(x_i)-p(x_j)}{p(x_j)-p(x_k)} = \frac{\sum_{\tau=i}^{j-1} d_{\tau}^2}{\sum_{\tau=j}^{k-1} d_{\tau}^2},
\end{equation}
where $1\leq i<j<k\leq T$. 
If we further constrain that $p(x_1)=+1$ and $p(x_T)=-1$ on this trajectory, $p(x_i)$ for points $\forall x_i \in X$ can be explicitly calculated as:
\begin{equation} \label{eq7}
    p(x_i) = \frac{\sum_{\tau=i}^{T-1} d_{\tau}^2-\sum_{\tau=1}^{i-1} d_{\tau}^2}{\sum_{\tau=1}^{T-1} d_{\tau}^2}.
\end{equation}

\begin{figure}[t!]
\centering
    \begin{subfigure}[b]{0.45\textwidth}
        \includegraphics[width=\textwidth]{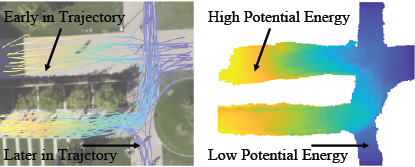}
        \vspace*{-0.5mm}
        \caption{Environmental potential field. Left: All the trajectories that traverse the image patch from left to right. Right: Potential field for the image patch. }
    \end{subfigure}~~~~~
    \begin{subfigure}[b]{0.45\textwidth}
        \includegraphics[width=\textwidth]{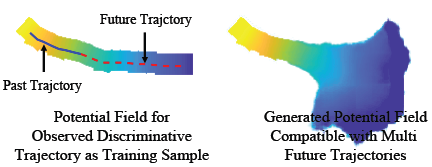}
        \vspace*{-0.5mm}
        \caption{Inertial potential field. Left: Training pair of past trajectory and the corresponding ground truth field. Right: Generated inertial potential field.}
    \end{subfigure}
   \caption{Environmental and inertial potential field generation. Best viewed in color.}
   \vspace*{-4mm}
   \label{fig:pf}
\end{figure}

We define that the trajectory $X$ resides in an image coordinate $(\mathcal{U},\mathcal{V})$, where $x_i = (u_i, v_i)$. We further generate a \textit{dense} potential field $\mathcal{P}_{X}\in \mathbb{R}^{H\times W}$ over $(\mathcal{U},\mathcal{V})$ given the sparse potential values on the trajectory and the pre-defined trajectory width $w$\footnote{The procedure is formally described in supplementary material (A3).}. 
An instance is shown in Figure~\ref{fig:pf}(b). 
In the rest of this paper, we use the notation $\mathcal{P}$ as the calculated ground truth potential field, and $\widehat{\mathcal{P}}$ as the estimated potential field from network inference.

\section{Potential Field Estimation}\label{stimuli_pf}
By introducing the potential field as representation, we transform the prediction problem to learning to generate a full potential field. The full potential field reflects the combined effects of all stimuli (environmental/inertial/social).
In this section, we demonstrate how we decompose the combined potential field into effects from the three stimuli. Such decomposition allows us to explicitly reason the formation with interpretablity of the target field. 

\noindent\textbf{Environmental potential field} 
Human motions obey the constraints from an environment, such as walking along the road and avoiding obstacles. 
To capture such constraints and eliminate the effect of individual preference, we gather a large number of trajectories that transverse the same image patch and train them together. 
Learning the potential field in a data driven way allows us to automatically “detect” multiple entrances and exits. It also solves the problem of combinatorial number of entrance-exit pairs, because only feasible combinations will survive in the potential field. 

\noindent\textbf{Inertial potential field} 
Human motion follows inertial force whose information is partially encoded in the history trajectory.
To capture such information and eliminate the influence of the environment, we collect trajectories over different environments and train the samples together. 
The resulting potential field works as a supplemental mechanism to environmental potential field. Thus, the future prediction will be dominated by inertial information when the environment can not provide sufficient constraints. Such cases include but are not limited to: (1) unstructured environments, (2) scenes with multiple exits, and (3) pedestrians who do not go to the exit directly. 

\noindent\textbf{Social force field} Without consideration of social context, pedestrian will move towards his/her intended destination with constraints from the environment. Social influence works as a force that makes the pedestrian deviate his/her original path to avoid collision or to maintain acceptable (social) distance with others. Other than generating a potential field, we explicitly model social force as a \textit{force} field, which will be demonstrated in details in Section~\ref{section_social} . 



\subsection{Environmental Potential Field}\label{3_2}
We define the input road structure image captured from a bird-eye view as $\mathcal{I} \in \mathbb{R}^{H \times W \times C}$, where $W$ and $H$ are the width and height of the image and $C$ is the number of channels. The observed trajectories on image patch $\mathcal{I}$ reflect environmental influence on human motions. Thus, we collect all agents' trajectories that traverse the scene from left to right as shown in Figure~\ref{fig:pf} (a). The trajectory set is denoted as $\mathbb{J}_\mathcal{I} = \{X^s \Big | s = 1,..,N\}$. 

We aim to learn a mapping from a bird-eye view road image $\mathcal{I}$ to an environmental potential field $\mathcal{P}_{\mathcal{I}}$ that captures agents' reaction within such physical surroundings, which can be formulated as:
\begin{equation} \label{eq_pi}
\widehat{\mathcal{P}}_\mathcal{I}=\kappa(\mathcal{I}, \Theta_{\mathcal{I}}),
\end{equation} 
where $\kappa(\cdot):\mathbb{R}^{H\times W\times C}\rightarrow \mathbb{R}^{H\times W}$ is a mapping function with $\Theta_{\mathcal{I}}$ being a set of trainable parameters. 
In this paper, we use an encoder-decoder structure~\cite{isola2017image} to model function $\kappa(\cdot)$.

To train the network, we treat each $(\mathcal{I}, \mathcal{P}_{X^s})$ as an input-output pair, where $\mathcal{P}_{X^s}$ is the ground truth potential field for trajectory $X^s$. By definition, the estimated $\widehat{\mathcal{P}}_\mathcal{I}$ should be compatible to each $\mathcal{P}_{X^s}$ at the region where $\mathcal{P}_{X^s}\neq 0$. 
The loss function for the network is thus given as:
\begin{equation}\label{eq:loss_I}
    \mathscr{L}_\mathcal{I} = \sum_{s=1}^N
    \norm{
    \mathcal{M}_{X^s}\cdot\left[\mathcal{P}_{X^s}-\widehat{\mathcal{P}}_\mathcal{I}\right]
    },
\end{equation}

\noindent where $\mathcal{M}_{X^s}\in\mathbb{R}^{H\times W}$ is a pixel-wise mask that acts as a indicator function for trajectory $X^s$ with $\mathcal{M}_{X^s}(u, v)=1 \text{~if~} \mathcal{P}_{X^s}(u,v)\neq 0$ and $\mathcal{M}_{X^s}(u, v)=\lambda$ otherwise. 
$\lambda<1$ is a weight parameter. 

The loss function enforces $\widehat{\mathcal{P}}_\mathcal{I}$ to be consistent with different field $\mathcal{P}_{X^s}$ in the scene, while having a regularization part $\lambda$ for the region where no one travels. In practice, we choose $\lambda=0.01$. Trajectories in the test set are not used during the process of estimating $\mathscr{L}_\mathcal{I}$ in the training phase.  
Note that our potential field is naturally shaped by the agents' trajectories, which encodes semantic traffic rules such as walking on the right side of the road. This is more representative to human behaviors than road structure segmentation~\cite{lee2017desire}.

The agents can transverse the region of interest from left to right, or from right to left. However, the potential fields for these two trajectories cancel out each other. To ensure the consistency of the potential field, we collect only the trajectories that travels from left to right during training. To utilize the whole trajectory dataset, we rotate each scene image (with the corresponding trajectories) 8 times with 45 degree each. For a certain trajectory in the training set, it transverses the scene from left to right in at least one rotated version of the original image patch. During the test phase, we crop the image scene centered at the agent’s current location $x_t$. The past trajectory of the agent is used to rotate trajectory and the corresponding image patch such that the agent is guaranteed to move from left to right\footnote{Illustration is shown in supplementary material (A4). }. Such pre-processing makes our environmental potential field to be \textbf{agent-specific} and \textbf{time-specific}, since the past motion at current time is used.

\subsection{Inertial Potential Field}\label{3_3}
For the evaluation of our method, we crop the full trajectory data $X$ into trajectory samples with length $n$ with observation length being $t$. We denote the observation part to be $X_{past}=\{x_\tau \Big | \tau = 1,...,t\}$ and the full trajectory sample to be $X_{sample}=\{x_\tau \Big | \tau = 1,...,n\}$\footnote{Following common practice, we use $t=8$ and $n=18~\text{or}~20$ in this paper. }. 

We aim to learn a mapping between the trajectory's past motion $X_{past}$ and its inertial potential field  $\mathcal{P}_{X_{sample}}$ that is compatible with the whole $n$ steps. It can be formulated as:

\begin{equation}
\widehat{\mathcal{P}}_{X_{sample}}=\phi(X_{past}, \Theta_X),
\end{equation}
where $\phi(\cdot)$ is a mapping function with $\Theta_X$ being a set of trainable parameters. 
The generated inertial potential field $\widehat{\mathcal{P}}_{X_{sample}}$ should show a distribution of possible future motions given the past trajectory. It corresponds to the multi-modal nature of future prediction (shown in Figure~\ref{fig:pf}(b)). To be consistent with environmental potential field $\widehat{\mathcal{P}}_{I}$ (Section~\ref{3_2}), the trajectory is translated and rotated accordingly. 

In practice, we use a neural network to model the function $\phi(\cdot)$, and treat $(X_{past}, \mathcal{P}_{X_{sample}})$ as an input-output pair. The loss function is then given by 
\begin{equation}
    \mathscr{L}_X = 
    \norm{
        \mathcal{M}_{X_{sample}}\cdot(\mathcal{P}_{X_{sample}}-\widehat{\mathcal{P}}_{X_{sample}})
    }
\end{equation}


\section{Trajectory Prediction}\label{sec:prediction}
In this section, we further demonstrate how the potential field representation can be used for trajectory prediction. 
We explicitly utilize the physics background embedded with the generated fields. This provides our method with strong logic proof and interpretability. 

With the generated potential field, the future trajectory can be calculated by ordinary differential equation. However, this step converts the potential field back to force, which overlooks the multi-modal nature of trajectory prediction. Due to the unpredictability of the future, road agents may have different trajectories even when their past motions are the same. To represent such unpredictability, we use two separate Gaussian distributions to model the target agent's motion direction and speed. Such methodology is also used and proved beneficial in other works in the literature~\cite{yang2019top,scholler2020constant}. 

We separate the pedestrian's velocity into motion direction (orientation) $\mathcal{O} \in \mathbb{R}^{H \times W\times 2}$ and speed $S \in \mathbb{R}^{n-t}$, where $n-t$ is the number of prediction frames. We model the distributions of motion direction and speed as Gaussians denoted by $\mathcal{N}(\mathcal{O}, \Sigma_\mathcal{O})$ and $\mathcal{N}(S, \Sigma_S)$ respectively. The predicted speed $S$, motion direction field $\mathcal{O}$ and social force field $\mathcal{F}\in \mathbb{R}^{H \times W\times 2}$ are fused to generate the displacement field $\mathcal{D}$. The final prediction result is given by recurrently moving current location to predicted location on field $\mathcal{D}$.

\subsection{Motion Field}\label{4_1}

Given environmental ($\mathcal{P}_{\mathcal{I}}$) and inertial potential field ($\mathcal{P}_X$), We learn a mapping from the potential fields to derive the corresponding direction field $\mathcal{O}_{\mathcal{I}}$ and $\mathcal{O}_X$ for future motion as:
\begin{align} \label{eq4_1}
    \mathcal{O}_{\mathcal{I}/X} &= \zeta(\widehat{\mathcal{P}}_{\mathcal{I}/X}, \Theta_{\mathcal{O}_{\mathcal{I}/X}}) 
\end{align}
where $\zeta(\cdot):\mathbb{R}^{H\times W}\rightarrow \mathbb{R}^{H\times W}$ with $\Theta_{\mathcal{O}}$ being the learnable parameters. $\mathcal{I}/X$ denotes $\mathcal{I}$ or $X$. $\mathcal{O}_X$ and $\mathcal{O}_\mathcal{I}$ are generated respectively. 

$\mathcal{O}_\mathcal{I}$ and $\mathcal{O}_X$ are the resulted motion directions of two independent stimuli on the target agent. 
We further merge the environmental and inertial motion fields into a single field. Following the additive property of force, we can thus fuse the two with a weighted sum by
\begin{equation}
    \mathcal{O}_{\mathcal{I}, X} = \mathcal{Y}\cdot\mathcal{O}_X + (1-\mathcal{Y})\cdot\mathcal{O}_\mathcal{I},
\end{equation}
where $\mathcal{Y}=\chi(\mathcal{O}_{X}, \mathcal{O}_{\mathcal{I}})$ is a pixel-wise weighting mask determined by the two motion fields. We drop $\mathcal{I}$ and $X$ for $\mathcal{O}$ in later sections.


In practice, we model the functions $\zeta(\cdot)$ (Eq.~\ref{eq4_1}) using neural networks.  In addition to mean values of the distributions, the networks also output the variance of the distributions. The ground truth of the motion direction is calculated from the trajectory data. The loss function then enforces the network to generate distributions of $\mathcal{N}(\mathcal{O}$,$\Sigma_{\mathcal{O}})$.
More specifically, we estimate the maximum likelihood of the ground truth samples given the generated mean and sigma values, and the loss is given by 
\begin{align}
    \mathscr{L}_\mathcal{O} = -\sum_\tau \log P\Big( \frac{v_\tau}{|v_\tau|} \Big| \mathcal{N}(\mathcal{O}(x_\tau), \Sigma_{\mathcal{O}}(x_\tau))\Big) 
\end{align}
where $v_{\tau}$ (in Eq.~\ref{eq2}) is the velocity of an agent at location $x_{\tau}$ at time $\tau$.


\subsection{Speed Prediction}\label{section_social}
Our observation is that the agent's future speed is encoded in inertial potential field $\widehat{\mathcal{P}}_{X_{sample}}$. On the other hand, environmental potential field do not carry such pedestrian specific information\footnote{The justification is provided in supplementary material (A5). }.

We learn a mapping from the estimated inertial potential field $\widehat{\mathcal{P}}_{X_{sample}}$ to the expected value of speed $S$ for the future motion, which is formulated as:
\begin{align} \label{eq4_2}
    S &= \psi(\widehat{\mathcal{P}}_{X_{sample}}, \Theta_\psi),
\end{align}
where $\psi(\cdot):\mathbb{R}^{H\times W}\rightarrow \mathbb{R}^{(n-t)}$, $n$ is the length of the whole trajectory and $t$ is the length of the past trajectory. 

In practice, we model the functions $\psi(\cdot)$ using neural networks.  The networks output the variance of the distributions in addition to the mean speed values. The maximum likelihood is estimated given the generated mean and sigma values, and the loss is:
\begin{align}
    \mathscr{L}_S =& -\sum_\tau \log P\Big(|v_\tau| \Big| \mathcal{N}(S(\tau), \Sigma_{S}(\tau))\Big)
\end{align}

\subsection{Social Force Field}\label{section_social}

We define the social force field $\mathcal{F}\in \mathbb{R}^{W\times H\times 2}$ to be the effect of other neighbor agents' influences on the target. More specifically, for each agent's coordinate location $(u,v)\in(\mathcal{U},\mathcal{V})$, we define $\mathcal{F}(u,v)\in \mathbb{R}^2$ as the vector that represents social pressure. Instances are shown in Figure~\ref{fig:qualitative_social}. 
For an agent $k$ with past trajectory $X_{past}^k = \{x_\tau^k \Big | \tau = 1,...,t\}$, we further define a set of neighbor's past trajectories as $\mathbb{J}_{near}^k = \{X_{past}^c \Big | \norm{x_t^c - x_t^k} \leq r\}$, where $t$ is the number of observation frames and $r$ is a pre-defined radius. 

We construct neighbors' field $\mathcal{B}_{\mathbb{J}_{near}^k}=\sum_c \mathcal{P}_{X^c_{past}}$ being the sum of potential fields from neighbors' past trajectory, and then learn a mapping from the neighbors' field $\mathcal{B}$ to the social force field $\mathcal{F}$ as:
\begin{align} \label{eq4_2}
    \mathcal{F} = \rho(\mathcal{B}_{\mathbb{J}_{near}^k}, \Theta_\rho),
\end{align}
where $\rho(\cdot):\mathbb{R}^{H\times W}\rightarrow \mathbb{R}^{H\times W\times 2}$, and $ \Theta_\rho$ is a set of trainable parameters\footnote{The supervision of the social field $\mathcal{F}$ is illustrated in supplementary material (A6). }. 

\subsection{Single and Multiple Future Prediction}\label{4_2}

For single future prediction, mean values $\mathcal{O}$ for motion direction, $S$ for speed, and social force field $\mathcal{F}$ are used to generate displacement field as follows:
\begin{equation}
    \mathcal{D}_{\tau} = \mathcal{O} \cdot S(\tau) + \mathcal{F},
\end{equation}
where $\mathcal{D}_{\tau}\in\mathbb{R}^{H\times W\times2}$ is a vector field with scale and $\tau \in \{t, t+1, ..., n-1\}$ is a set of desired prediction time. The displacement field set $\mathscr{D}=\{\mathcal{D}_{t},\mathcal{D}_{t+2},...,\mathcal{D}_{n-1}\}$ provides the complete motion of each pixel at every time step. Then, the trajectory prediction is given by recurrently moving and updating the previous location $x_{\tau}$  by 
\begin{equation}\label{Eq_disp}
    x_{\tau+1} = x_{\tau} + \mathcal{D}_{\tau}(x_{\tau}),
\end{equation}
where $t \leq \tau < n$. 

For multi-modal future prediction, we sample instances $\mathcal{O}^j$ for motion direction and $S^j$ for speed from the distribution $\mathcal{N}(\mathcal{O}, \Sigma_\mathcal{O})$ and $\mathcal{N}(S, \Sigma_S)$, respectively, and combine with social force field $\mathcal{F}$ to generate displacement field
$\mathcal{D}_{\tau}^j = \mathcal{O}^j \cdot \mathcal{S}^j(\tau) + \mathcal{F},$ 
where $1\leq j\leq K$ is the prediction index and $K$ is the number of generated predictions. 
The predicted trajectory is then generated by recurrently applying $\mathcal{D}_{\tau}^j$ from the previous location $x_{\tau^j}$ by
$x_{\tau+1}^j = x_{\tau}^j + \mathcal{D}^j_{\tau}(x_{\tau^j})$.
Note that the predicted trajectories for both single- and multi-modal prediction are generated from previously learned fields with \textit{no extra parameters}. In practice\footnote{The details are illustrated in supplementary material (A7). }, we use spatial transformer layer~\cite{jaderberg2015spatial} to achieve Eq.~\ref{Eq_disp}. 


\begin{table*}[t!]
\centering
\begin{tabular}{c||c|c|c|c|c||c}
\hline
 Method & ETH & Hotel & Univ & Zara01 & Zara02 & Average \\ \hline\hline
 Lin & 1.33/2.94 & 0.39/0.72 & 0.82/1.59 & 0.62/1.21 & 0.77/1.48 & 0.79/1.59 \\
 S-LSTM~\cite{alahi2016social} & 1.09/2.35 & 0.79/1.76 & 0.67/1.40 & 0.47/1.00 & 0.56/1.17 & 0.72/1.54 \\
 PIF~\cite{liang2019peeking} & 0.88/1.98 & 0.36/0.74  & \textbf{0.62}/\textbf{1.32} & \textbf{0.42}/\textbf{0.90} & \textbf{0.34}/0.75 & 0.52/1.14 \\ \hline
 Ours wo social & 0.96/1.91 & 0.30/0.55& 0.70/1.49 & 0.51/1.17 & 0.39/0.87 & 0.57/1.20 \\
 Ours & \textbf{0.85}/\textbf{1.66} & \textbf{0.22}/\textbf{0.42}  & 0.64/1.37 & 0.44/0.97 & 0.36/\textbf{0.73} & \textbf{0.50}/\textbf{1.03} \\ \hline\hline
 S-GAN-P~\cite{gupta2018social} & 0.87/1.62 & 0.67/1.37 & 0.76/1.52 & 0.35/0.68 & 0.42/0.84 &  0.61/1.21 \\
 Sophie~\cite{Sadeghian_2019_CVPR} & 0.70/1.43 & 0.76/1.67 & \textbf{0.54}/1.24 & 0.30/0.63 & 0.38/0.78 & 0.54/1.15 \\ 
 S-BiGAT~\cite{kosaraju2019social} & \textbf{0.69}/\textbf{1.29} & 0.49/1.01 & 0.55/1.32 & \textbf{0.30}/\textbf{0.62} & 0.36/0.75 & 0.48/1.00 \\
 PIF-20~\cite{liang2019peeking} & 0.73/1.65 & 0.30/0.59 & 0.60/1.27 & 0.38/0.81 & \textbf{0.31}/0.68 & 0.46/1.00 \\ \hline
 Ours & 0.79/1.49 & \textbf{0.22}/\textbf{0.38} & 0.58/\textbf{1.19} & 0.36/0.75 & \textbf{0.31}/\textbf{0.63} & \textbf{0.45}/\textbf{0.89} \\ \hline\hline
\end{tabular}
\caption{Qualitative results on ETH/UCY dataset for both single and multi prediction. ADE and FDE are reported in meters. }
\label{table:ETH_UCY}
\vspace*{-4mm}
\end{table*}

\begin{table*}[t!]
\centering
\begin{tabular}{c|c|cccc}
\hline
Category & Method   & 1.0 sec   & 2.0 sec   & 3.0 sec     & 4.0 sec     \\ \hline\hline
& Linear & ~~-~~/2.58 & ~~-~~/5.37 & ~~-~~/8.74 & ~~-~~/12.54  \\
Single Prediction & S-LSTM~\cite{alahi2016social}   & 1.93/3.38 & 3.24/5.33 & 4.89/9.58 & 6.97/14.57  \\
(State-of-the-art) & DESIRE~\cite{lee2017desire}   & ~~-~~/2.00 & ~~-~~/4.41 & ~~-~~/7.18 & ~~-~~/10.23 \\
& Gated-RN~\cite{Choi_2019_ICCV} & 1.71/2.23 & 2.57/3.95 & 3.52/6.13 & 4.60/8.79~~ \\ \hline
Single Prediction & Inertial & 0.91/1.39 &  1.81/3.29 & 2.84/5.75 & 4.08/8.61~~ \\
(Ours) & Inertial + Env. & 0.74/1.17 &  1.57/2.95 & 2.56/5.36 & 3.77/8.28~~ \\
& Ours Full & \textbf{0.73}/\textbf{1.16} & \textbf{1.55}/\textbf{2.89} & \textbf{2.51}/\textbf{5.25} & \textbf{3.70}/\textbf{8.10}~~ \\ \hline \hline
Multi Prediction & CVAE~\cite{sohn2015learning}   & 1.84    & 3.93    & 6.47      & 9.65     \\ 
(State-of-the-art) & DESIRE~\cite{lee2017desire}   & 1.29    & 2.35    & \textbf{3.47}      & 5.33     \\ \hline
Multi (Ours) & Ours & \textbf{1.10} & \textbf{2.33} & 3.62 & \textbf{4.92}\\  \hline
\end{tabular}
\caption{Quantitative results on SDD dataset for both single and multi prediction. ADE and FDE are reported in pixel coordinates at 1/5 resolution as proposed in~\cite{lee2017desire}. Our method outperforms baselines consistently. }
\label{table:SDD}
\vspace*{-4mm}
\end{table*}

\section{Experiments}
The whole pipeline is implemented using deep neural networks and the details are presented in supplementary material (A8). We evaluate our algorithm on three widely used benchmark datasets ETH~\cite{pellegrini2009you} / UCY~\cite{lerner2007crowds} and Stanford Drone Dataset (SDD)~\cite{robicquet2016learning}. All datasets contain annotated trajectories of real world humans. The ETH / UCY dataset has 5 scenes, while the SDD dataset has eight scenes of 60 videos with more complex road structures. 
\subsection{Quantitative Results}

We quantitatively evaluate our method on both ETH/UCY and SDD dataset, and compare our model with the state-of-the-art methods. 

For the ETH/UCY dataset, we adopt the same experimental setting of~\cite{alahi2016social,gupta2018social,Sadeghian_2019_CVPR,liang2019peeking}, where we split the trajectories into 8 seconds segments, and use 3.2 second for observations and 4.8 seconds for prediction and evaluation. We use four scenes for training and the remaining scene for testing in a leave-one-out cross-validation fashion to test the method's ability to generalize to unseen scenes. Average displacement error (ADE) and final displacement error (FDE) are reported in meters for 4.8s in the future.

Table~\ref{table:ETH_UCY} shows the quantitative comparison for single/discriminative prediction (upper part of the table) with linear, Social LSTM~\cite{alahi2016social}, and Peeking into the Future~\cite{liang2019peeking}. We also show ablation test for our model with and without social field. Our method outperforms previous methods in most scenes. In addition, we evaluate our performance on multi/generative prediction with social GAN~\cite{gupta2018social}, Sophie~\cite{Sadeghian_2019_CVPR}, Social BiGAT~\cite{kosaraju2019social} and Peeking into the Future~\cite{liang2019peeking} in bottom part of table~\ref{table:ETH_UCY}. $K=20$ possible trajectories are used for evaluation. 

\begin{figure*}[t!]
\centering
    \includegraphics[width=\textwidth]{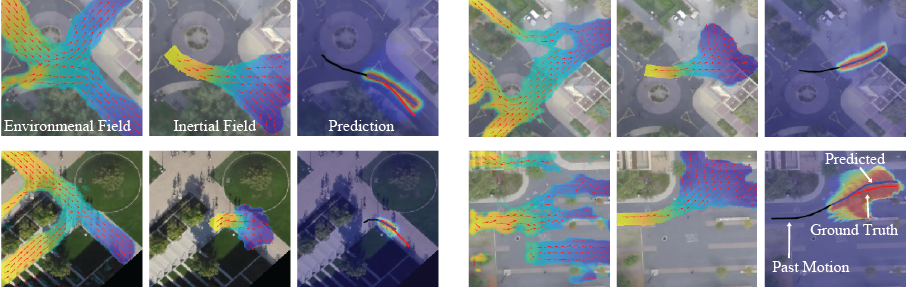}
    \caption{Qualitative results. For each instance, we show: (1) Environmental field. (2) Inertial field. (3) Final prediction results. We show the future distribution heatmap of the target agent, with red and blue denoting high and low probability. Left: Our potential field is able to recognize complicated road structure (roundabout/crossing) and generate reasonable motion field. Right: Our model is able to predict turning according to the context. Best viewed in color.}
    \vspace*{-4mm}
    \label{fig:qualitative}
\end{figure*}

For SDD, we randomly divide 60 videos into training and testing set with a ratio of 4:1, which is similar to~\cite{Choi_2019_ICCV,lee2017desire}. Since different videos shoot different environments, such standard split also provide evidence on our method's adaptive ability. We split the trajectories into 7.2 seconds segments, and use 3.2 seconds for observation and 4 seconds for prediction and evaluation. Raw RGB road images are used with no additional annotation and pre-processing except cropping and rotation. ADE and FDE are reported for 1s, 2s, 3s and 4s in future. The errors are reported in pixel coordinates in 1/5 resolution. 

Table~\ref{table:SDD} shows the quantitative comparison for single-modal prediction and multi-modal prediction. For single-modal prediction, we compare our model with S-LSTM~\cite{alahi2016social}, DESIRE~\cite{lee2017desire}, and Gated-RN~\cite{Choi_2019_ICCV}. We also provide ablation test of our model by showing the results with only inertial field, with inertial and environmental field, and our full modal with inertial/environmental/social fields together. Our method outperforms previous methods with only inertial stimuli. We additionally compare our multi-modal prediction with CVAE~\cite{sohn2015learning} and DESIRE~\cite{lee2017desire}, and report FDE. We predict $K=20$ trajectories for evaluation.

The quantitative evaluations prove that our method can robustly handle the future motion prediction in everyday activities which include unstructured environments (as ETH/UCY captures human motion in relative open area), environments with multiple entrances and exits (as SDD dataset contains environments with complex topology), and situations that social interaction is present (as both the datasets captured human activities in crowded scenes). By outperforming the state-of-the-art methods, our method demonstrates that intermediate supervision using interpretable representations is beneficial for information extraction. The ablation test further illustrate that adding road structure and social context into the pipeline is beneficial, and the proposed pipeline is efficient in extracting the information. It further validates our usage of unified representations to merge different domain knowledge.The improvements between single-modal prediction and multi-modal prediction show that the generated distributions capture the unpredictability of the future. 

\subsection{Qualitative Results}
\begin{figure}[t!]
\centering
    \includegraphics[width=\textwidth]{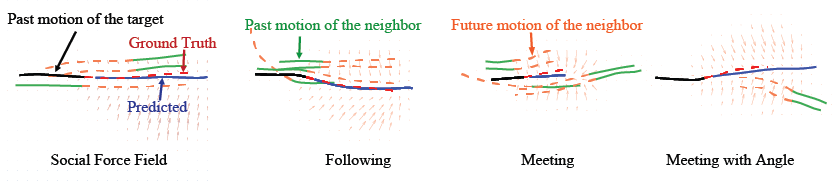}
    \caption{Social behavior. Social force field is generated based on the neighbors' motion history and influences the final prediction. We show the algorithm's reaction with scenarios such as following, meeting and meeting with an angle. }
    \vspace*{-4mm}
    \label{fig:qualitative_social}
\end{figure}

We qualitatively evaluate our method in Figure~\ref{fig:qualitative}. It shows that our model can deal with different challenging road structures (open area/straight road/ crossing/roundabout) and diverse motions (standing still/going straight/taking turn). As shown on the top right case, our potential field not only gives walkable area, but also learns walking habit of humans (walking on the right side of the road) automatically in a data-driven manner. Such information cannot be obtained from other labels such as road segmentation. The information from environmental and inertial information can be merged reasonably and compensate each other to generate plausible future trajectories. We also demonstrate that our method can deal with interaction intensive scenarios such as in Figure~\ref{fig:qualitative_social}, which shows the context of following, meeting and meeting with an angle. We provide more qualitative evaluations in the supplementary material.

\section{Conclusion}
Predicting future motion of road agents is a crucial and challenging task. We propose to use potential field as an interpretable and unified representation for human trajectory prediction. This enables us to not only fuse the information of different stimuli more reasonably, but also allows to supervise and evaluate the intermediate learning progress of neural networks. Potential/force fields are generated to represent the effect of the environmental force, inertial force, and social force on the target agent. We further estimate future velocity direction and magnitude from potential fields, which are modeled as Gaussian distributions to account for the unpredictability of the future. The predicted future trajectory is generated by recurrently moving past location on the displacement field. We test our model on three challenging benchmark datasets. The results show that our method can deal with complicated context while achieving state-of-the-art performances. 

\clearpage
%
%
\bibliographystyle{splncs04}
\bibliography{egbib}

\begin{thebibliography}{10}
\providecommand{\url}[1]{\texttt{#1}}
\providecommand{\urlprefix}{URL }
\providecommand{\doi}[1]{https://doi.org/#1}

\bibitem{alahi2016social}
Alahi, A., Goel, K., Ramanathan, V., Robicquet, A., Fei-Fei, L., Savarese, S.:
  Social lstm: Human trajectory prediction in crowded spaces. In: Proceedings
  of the IEEE conference on computer vision and pattern recognition. pp.
  961--971 (2016)

\bibitem{barraquand1992numerical}
Barraquand, J., Langlois, B., Latombe, J.C.: Numerical potential field
  techniques for robot path planning. IEEE transactions on systems, man, and
  cybernetics  \textbf{22}(2),  224--241 (1992)

\bibitem{chaiklin2003zone}
Chaiklin, S.: The zone of proximal development in vygotsky’s analysis of
  learning and instruction. Vygotsky’s educational theory in cultural context
   \textbf{1},  39--64 (2003)

\bibitem{chandra2019traphic}
Chandra, R., Bhattacharya, U., Bera, A., Manocha, D.: Traphic: Trajectory
  prediction in dense and heterogeneous traffic using weighted interactions.
  In: Proceedings of the IEEE Conference on Computer Vision and Pattern
  Recognition. pp. 8483--8492 (2019)

\bibitem{Choi_2019_ICCV}
Choi, C., Dariush, B.: Looking to relations for future trajectory forecast. In:
  Proceedings of the IEEE International Conference on Computer Vision (ICCV)
  (October 2019)

\bibitem{ge2002dynamic}
Ge, S.S., Cui, Y.J.: Dynamic motion planning for mobile robots using potential
  field method. Autonomous robots  \textbf{13}(3),  207--222 (2002)

\bibitem{griffiths2005introduction}
Griffiths, D.J.: Introduction to electrodynamics (2005)

\bibitem{gupta2018social}
Gupta, A., Johnson, J., Fei-Fei, L., Savarese, S., Alahi, A.: Social gan:
  Socially acceptable trajectories with generative adversarial networks. In:
  Proceedings of the IEEE Conference on Computer Vision and Pattern
  Recognition. pp. 2255--2264 (2018)

\bibitem{hasan2018mx}
Hasan, I., Setti, F., Tsesmelis, T., Del~Bue, A., Galasso, F., Cristani, M.:
  Mx-lstm: mixing tracklets and vislets to jointly forecast trajectories and
  head poses. In: Proceedings of the IEEE Conference on Computer Vision and
  Pattern Recognition. pp. 6067--6076 (2018)

\bibitem{helbing1995social}
Helbing, D., Molnar, P.: Social force model for pedestrian dynamics. Physical
  review E  \textbf{51}(5), ~4282 (1995)

\bibitem{hwang1992potential}
Hwang, Y.K., Ahuja, N.: A potential field approach to path planning. IEEE
  Transactions on Robotics and Automation  \textbf{8}(1),  23--32 (1992)

\bibitem{isola2017image}
Isola, P., Zhu, J.Y., Zhou, T., Efros, A.A.: Image-to-image translation with
  conditional adversarial networks. In: Proceedings of the IEEE conference on
  computer vision and pattern recognition. pp. 1125--1134 (2017)

\bibitem{jaderberg2015spatial}
Jaderberg, M., Simonyan, K., Zisserman, A., et~al.: Spatial transformer
  networks. In: Advances in neural information processing systems. pp.
  2017--2025 (2015)

\bibitem{kalman1960new}
Kalman, R.E.: A new approach to linear filtering and prediction problems
  (1960)

\bibitem{kosaraju2019social}
Kosaraju, V., Sadeghian, A., Mart{\'\i}n-Mart{\'\i}n, R., Reid, I.,
  Rezatofighi, H., Savarese, S.: Social-bigat: Multimodal trajectory
  forecasting using bicycle-gan and graph attention networks. In: Advances in
  Neural Information Processing Systems. pp. 137--146 (2019)

\bibitem{lake2017building}
Lake, B.M., Ullman, T.D., Tenenbaum, J.B., Gershman, S.J.: Building machines
  that learn and think like people. Behavioral and brain sciences  \textbf{40}
  (2017)

\bibitem{lee2017desire}
Lee, N., Choi, W., Vernaza, P., Choy, C.B., Torr, P.H., Chandraker, M.: Desire:
  Distant future prediction in dynamic scenes with interacting agents. In:
  Proceedings of the IEEE Conference on Computer Vision and Pattern
  Recognition. pp. 336--345 (2017)

\bibitem{lerner2007crowds}
Lerner, A., Chrysanthou, Y., Lischinski, D.: Crowds by example. In: Computer
  graphics forum. vol.~26, pp. 655--664. Wiley Online Library (2007)

\bibitem{liang2019peeking}
Liang, J., Jiang, L., Niebles, J.C., Hauptmann, A.G., Fei-Fei, L.: Peeking into
  the future: Predicting future person activities and locations in videos. In:
  Proceedings of the IEEE Conference on Computer Vision and Pattern
  Recognition. pp. 5725--5734 (2019)

\bibitem{luber2010people}
Luber, M., Stork, J.A., Tipaldi, G.D., Arras, K.O.: People tracking with human
  motion predictions from social forces. In: 2010 IEEE International Conference
  on Robotics and Automation. pp. 464--469. IEEE (2010)

\bibitem{pellegrini2009you}
Pellegrini, S., Ess, A., Schindler, K., Van~Gool, L.: You'll never walk alone:
  Modeling social behavior for multi-target tracking. In: 2009 IEEE 12th
  International Conference on Computer Vision. pp. 261--268. IEEE (2009)

\bibitem{reif1999social}
Reif, J.H., Wang, H.: Social potential fields: A distributed behavioral control
  for autonomous robots. Robotics and Autonomous Systems  \textbf{27}(3),
  171--194 (1999)

\bibitem{robicquet2016learning}
Robicquet, A., Sadeghian, A., Alahi, A., Savarese, S.: Learning social
  etiquette: Human trajectory understanding in crowded scenes. In: European
  conference on computer vision. pp. 549--565. Springer (2016)

\bibitem{1905.06113}
Rudenko, A., Palmieri, L., Herman, M., Kitani, K.M., Gavrila, D.M., Arras,
  K.O.: Human motion trajectory prediction: A survey (2019)

\bibitem{Sadeghian_2019_CVPR}
Sadeghian, A., Kosaraju, V., Sadeghian, A., Hirose, N., Rezatofighi, H.,
  Savarese, S.: Sophie: An attentive gan for predicting paths compliant to
  social and physical constraints. In: The IEEE Conference on Computer Vision
  and Pattern Recognition (CVPR) (June 2019)

\bibitem{scholler2020constant}
Sch{\"o}ller, C., Aravantinos, V., Lay, F., Knoll, A.: What the constant
  velocity model can teach us about pedestrian motion prediction. IEEE Robotics
  and Automation Letters  (2020)

\bibitem{sohn2015learning}
Sohn, K., Lee, H., Yan, X.: Learning structured output representation using
  deep conditional generative models. In: Advances in neural information
  processing systems. pp. 3483--3491 (2015)

\bibitem{su2017predicting}
Su, S., Pyo~Hong, J., Shi, J., Soo~Park, H.: Predicting behaviors of basketball
  players from first person videos. In: Proceedings of the IEEE Conference on
  Computer Vision and Pattern Recognition. pp. 1501--1510 (2017)

\bibitem{wang2007gaussian}
Wang, J.M., Fleet, D.J., Hertzmann, A.: Gaussian process dynamical models for
  human motion. IEEE transactions on pattern analysis and machine intelligence
  \textbf{30}(2),  283--298 (2007)

\bibitem{williams1998prediction}
Williams, C.K.: Prediction with gaussian processes: From linear regression to
  linear prediction and beyond. In: Learning in graphical models, pp. 599--621.
  Springer (1998)

\bibitem{xue2018ss}
Xue, H., Huynh, D.Q., Reynolds, M.: Ss-lstm: A hierarchical lstm model for
  pedestrian trajectory prediction. In: 2018 IEEE Winter Conference on
  Applications of Computer Vision (WACV). pp. 1186--1194. IEEE (2018)

\bibitem{yagi2018future}
Yagi, T., Mangalam, K., Yonetani, R., Sato, Y.: Future person localization in
  first-person videos. In: Proceedings of the IEEE Conference on Computer
  Vision and Pattern Recognition. pp. 7593--7602 (2018)

\bibitem{yang2019top}
Yang, D., Li, L., Redmill, K., {\"O}zg{\"u}ner, {\"U}.: Top-view trajectories:
  A pedestrian dataset of vehicle-crowd interaction from controlled experiments
  and crowded campus. In: 2019 IEEE Intelligent Vehicles Symposium (IV). IEEE
  (2019)

\bibitem{zanlungo2011social}
Zanlungo, F., Ikeda, T., Kanda, T.: Social force model with explicit collision
  prediction. EPL (Europhysics Letters)  \textbf{93}(6),  68005 (2011)

\bibitem{Zhao_2019_CVPR}
Zhao, T., Xu, Y., Monfort, M., Choi, W., Baker, C., Zhao, Y., Wang, Y., Wu,
  Y.N.: Multi-agent tensor fusion for contextual trajectory prediction. In: The
  IEEE Conference on Computer Vision and Pattern Recognition (CVPR) (June 2019)

\end{thebibliography}
\end{document}